\documentclass[10pt,twocolumn,letterpaper]{article}

\usepackage{iccv}
\usepackage{times}
\usepackage{epsfig}
\usepackage{graphicx}
\usepackage{amsmath}
\usepackage{amssymb}

\usepackage{booktabs}
\usepackage{multirow}
\usepackage{algorithm,algorithmic}

\usepackage[pagebackref=true,breaklinks=true,letterpaper=true,colorlinks,bookmarks=false]{hyperref}

\iccvfinalcopy 

\def\iccvPaperID{7472} 

\ificcvfinal\fi

\begin{document}

\title{MRN: Multiplexed Routing Network \\for Incremental Multilingual Text Recognition}

\author{
Tianlun Zheng$^{1,2}$,  Zhineng Chen$^{1,2}$\thanks{Corresponding Author.},  Bingchen Huang$^{1,2}$,  Wei Zhang$^3$ and Yu-Gang Jiang$^{1,2}$\\
$^1$School of Computer Science, Fudan University, China\\
$^2$Shanghai Collaborative Innovation Center of Intelligent Visual Computing, China\\
$^3$Gaoding AI, China\\
{\tt\small\{tlzheng21, bchuang21\}@m.fudan.edu.cn,
\{zhinchen, ygj\}@fudan.edu.cn,
wzhang.cu@gmail.com}
}

\maketitle
\ificcvfinal\thispagestyle{empty}\fi

\maketitle

\begin{abstract}
Multilingual text recognition (MLTR) systems typically focus on a fixed set of languages, which makes it difficult to handle newly added languages or adapt to ever-changing data distribution. In this paper, we propose the Incremental MLTR (IMLTR) task in the context of incremental learning (IL), where different languages are introduced in batches. IMLTR is particularly challenging due to rehearsal-imbalance, which refers to the uneven distribution of sample characters in the rehearsal set, used to retain a small amount of old data as past memories. To address this issue, we propose a Multiplexed Routing Network (MRN). MRN trains a recognizer for each language that is currently seen. Subsequently, a language domain predictor is learned based on the rehearsal set to weigh the recognizers. Since the recognizers are derived from the original data, MRN effectively reduces the reliance on older data and better fights against catastrophic forgetting, the core issue in IL. We extensively evaluate MRN on MLT17 and MLT19 datasets. It outperforms existing general-purpose IL methods by large margins, with average accuracy improvements ranging from 10.3\% to 35.8\% under different settings. Code is available at \url{https://github.com/simplify23/MRN}.

\end{abstract}

\section{Introduction}
Scene text recognition (STR) is a task aiming to read text in natural scenes. Recent advances in deep learning have significantly improved the accuracy of STR, allowing it to recognize text in the presence of font variations, distortions, and noise interference \cite{ShiBY17crnn,shi2018aster,wang2021visionlan,sheng2019nrtr,fang2022abinet++,zheng2023tps++}. As countries and cultures are more interconnected, the task of simultaneously recognizing multiple languages, i.e., multilingual text recognition (MLTR), has also become more important.
Existing methods typically address this challenge by training on mixed multilingual data \cite{buvsta2018e2emlt,baek2020crafts,nayef2019mlt2019} or designing independent language blocks \cite{huang2021multiocr,fujii2017seqscript,gomez2017improving}. However, when each time a new language is added, the above methods need retraining on a dataset mixing the old and new languages. This increases the training cost \cite{Rebuffi2017icarl,Yan2021DER} and also may lead to an imbalance \cite{belouadah2019IL2M,delange2022clsurvey} between old and new data.

\begin{figure}[]
\centering
\includegraphics[width=0.45\textwidth]{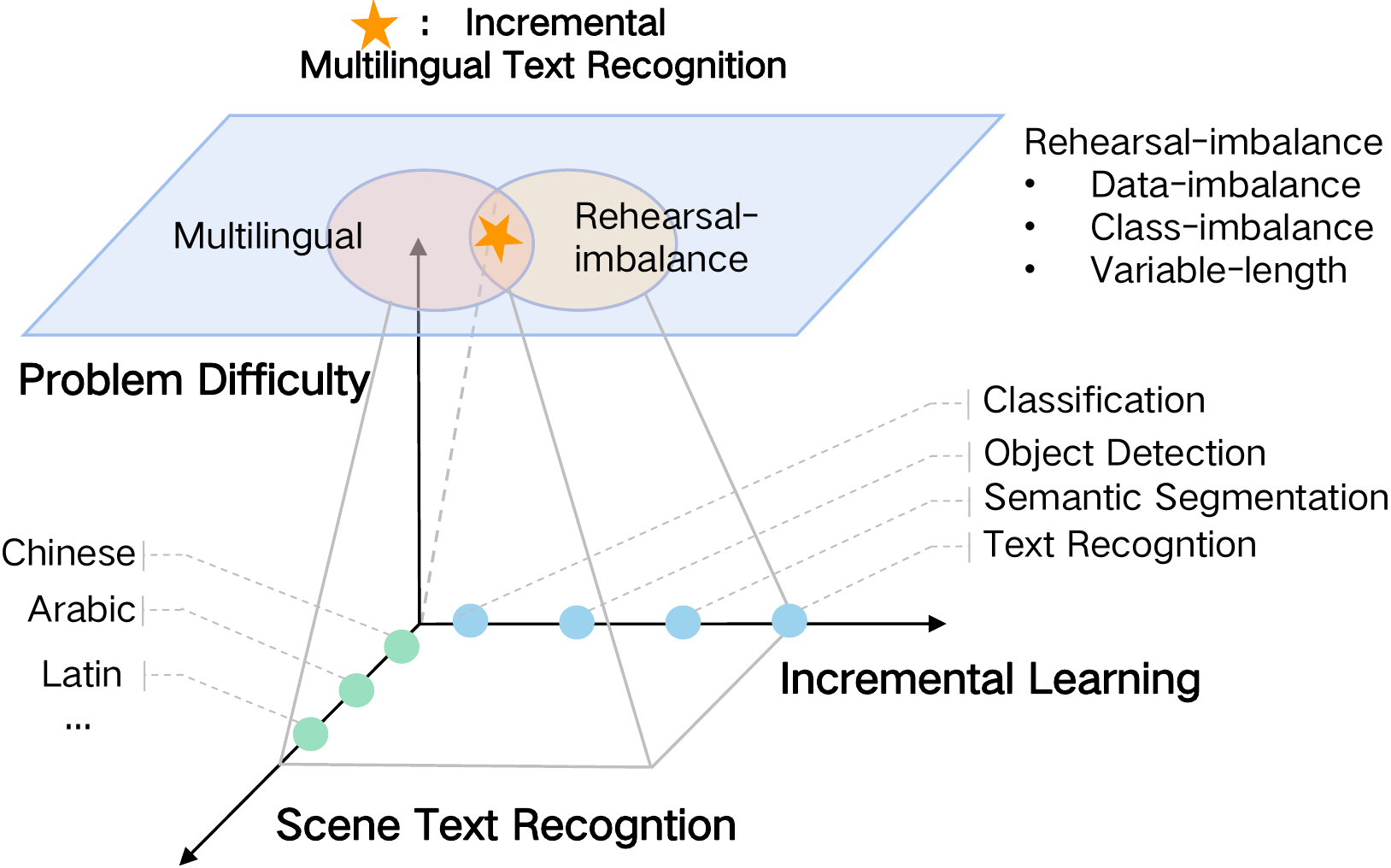} 
\caption{
Incremental multilingual text recognition (IMLTR) focuses on the practical scenario where different languages are introduced sequentially. The goal is to accurately recognize the newly introduced language while maintaining high recognition accuracy for previously seen languages.
IMLTR introduces a task focusing on text recognition that faces rehearsal-imbalance challenges.
}
\label{fig1:motivation}
\end{figure}

Incremental learning (IL) is designed for scenarios where new data is continuously learned and typically, the old samples are maintained by a small ratio. The collection of old samples is referred to as the rehearsal set \cite{zhang2022rcil,hu2021distilling}, which serves as limited past memories. IL aims to learn the new data well while minimizing forgetting the past learned knowledge. Most existing studies \cite{Rebuffi2017icarl, belouadah2019IL2M, zhao2020wa,huang2023resolving} conduct experiments on balanced datasets and maintain a constant number of classes at each learning step. 
However, in real-world scenarios, the number of classes and samples may differ across steps, leading to imbalanced datasets. To address these issues, IL2M \cite{belouadah2019IL2M} alleviated class-imbalance by storing statistics of old classes rather than samples. Delange et al. De Lange et al. \cite{delange2022clsurvey} surveyed typical IL methods on datasets and solutions with different data imbalances. Despite progress made, research on data and class imbalance is still in its infancy stage. Moreover, as illustrated in Fig.~\ref{fig1:motivation}, there is currently no research introducing IL to STR.



We rewrite MLTR in the context of IL. Languages are treated as tasks and characters are their classes. During training, the model only observes the newly arrived language data and a small amount of data from old languages. The recognition model is expected to maintain the ability to recognize characters of all languages that it has encountered before, regardless of whether their data are still available or discarded. We term this problem incremental multilingual text recognition (IMLTR).

\begin{figure}[]
\centering
\includegraphics[width=0.48\textwidth]{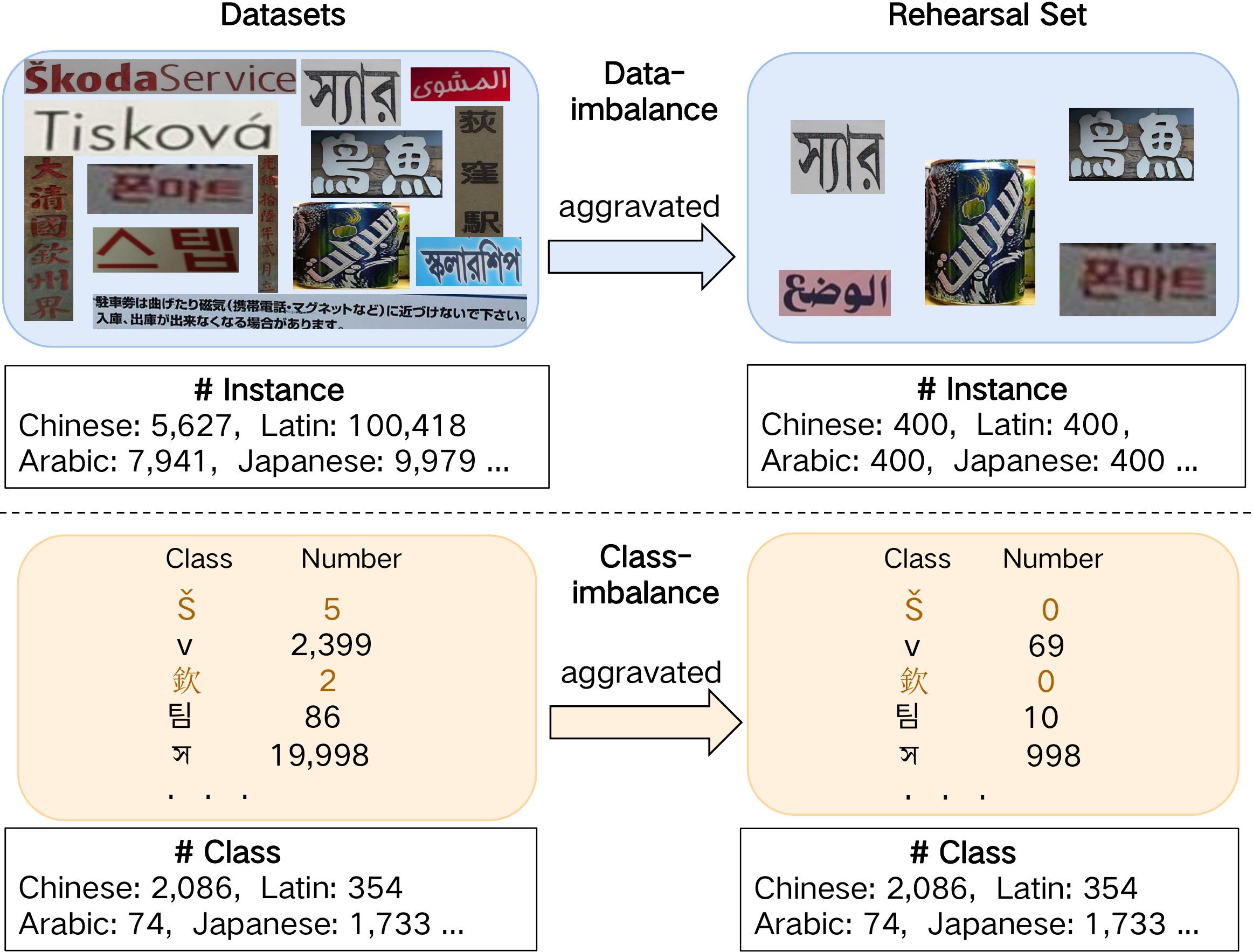} 
\caption{The showcase of rehearsal-imbalance. Data-imbalance (top) and class-imbalance (bottom) are severely aggravated from the full dataset to the rehearsal set, while the character classes to be read remain the same, making IMLTR particularly challenging.}
\label{fig2:rehearsal-imbalance}
\end{figure}

IMLTR poses significant challenges to IL approaches due to its unbalanced features. 1) At the dataset level, it is difficult to collect sufficient training data for minority languages such as Bangla compared to popular languages such as English and Chinese, which affects the quality of recognition models. 2) At the language level, the size of character sets varies from tens to thousands across different languages, which leads to data imbalance. 3) At the character level, the occurrence frequency of characters follows a long-tailed distribution, leading to class imbalance. In addition, IMLTR faces the problem of variable length recognition, where text instances are the recognizing unit instead of character classes. Therefore, IL methods cannot sample characters as evenly as required in the context of IMLTR, resulting in a significant fraction of characters not being included in the rehearsal data, as shown in Fig.~\ref{fig2:rehearsal-imbalance}. This phenomenon is summarized as rehearsal-imbalance in Fig.~\ref{fig1:motivation}. Rehearsal-imbalance leads to catastrophic forgetting, where forgotten characters cannot be recognized. Therefore, there is an urgent need to develop new methods to overcome it.

Although the rehearsal set does not ensure full coverage of all interlingual character classes, it is still adequate for training a language domain predictor to identify the languages. Motivated by this observation, we propose a novel Multiplexed Routing Network (MRN) for IMLTR. MRN involves training a new text recognition model at each learning step and utilizing it and previously trained models for parallel feature extraction. A domain MLP router is designed to receive these features and predict the probability over the languages. Meanwhile, these features are used for character recognition in their own domain by feeding them to the multi-lingual modeling module. Finally, we fuse the results obtained at both the language domain and character levels to decode the recognized character sequence.

Our contributions can be summarized as follows. First, we introduce the IMLTR task, the first effort to adapt IL to text recognition. It contributes to the exploration of other practical scenarios for text recognition. Second, we develop MRN to address the rehearsal-imbalance problem in ILMTR. It is a dynamic and scalable architecture that is compatible with various IL methods and recognition models. Third, experiments on two benchmarks show that MRN significantly outperforms existing general-purpose IL methods, achieving accuracy improvements ranging from 10.3\% to 27.4\% under different settings.

\begin{figure*}[]
\centering
\includegraphics[width=1\textwidth]{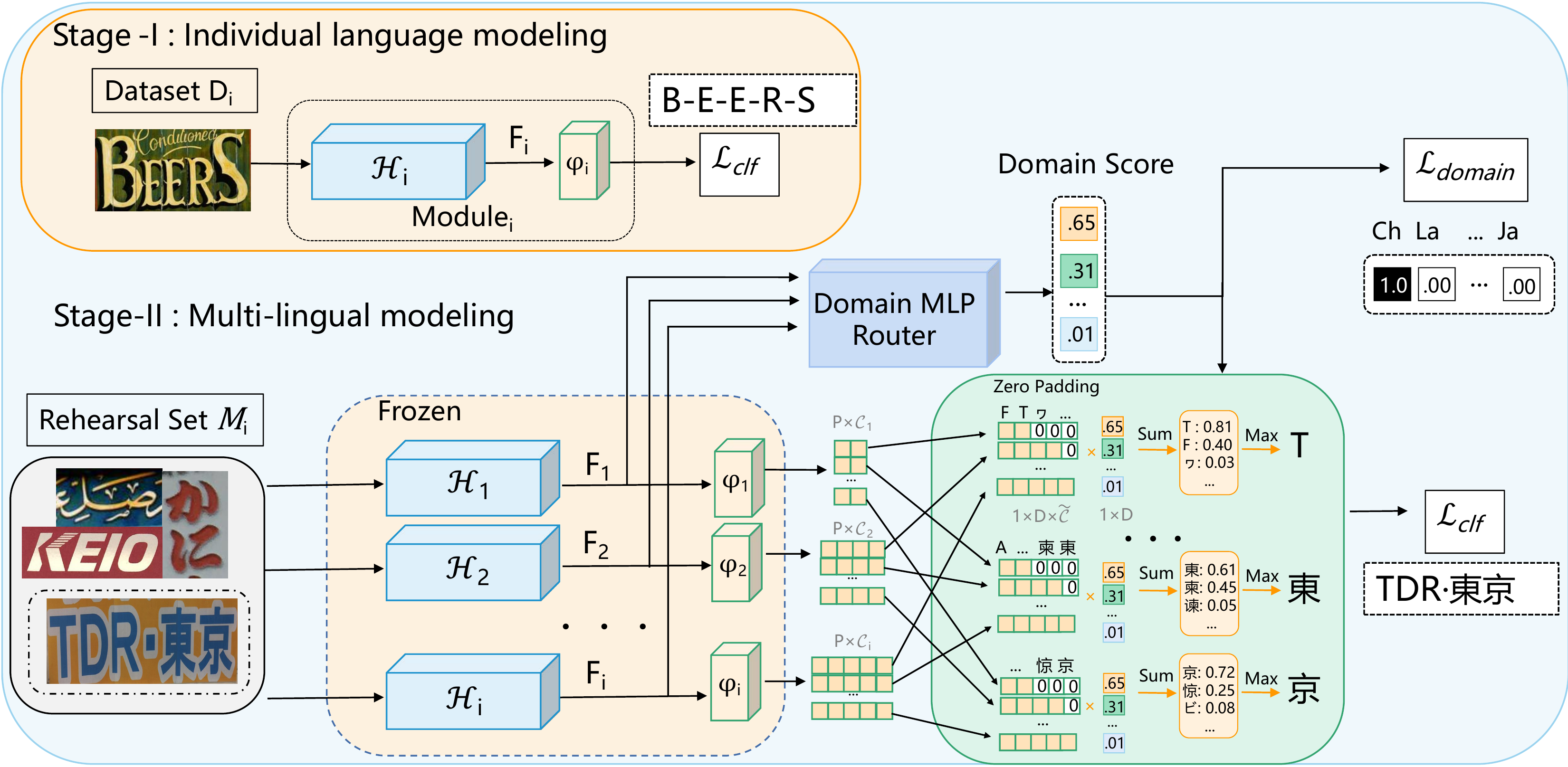} 
\caption{An overview of MRN. In stage-I, text recognizers are trained language-by-language. While in stage-II, these recognizers are frozen for feature extraction. The Domain MLP Router, which is trained based on the rehearsal set, is proposed to predict the likelihood distribution over the languages. Meanwhile, a padded classification layer is constructed, where the parallel predicted text sequences and likelihood distributions are merged to generate the decoded character sequence.}
\label{fig:overview}
\end{figure*}
\section{Related Work}
\subsection{Incremental Learning (IL)}
IL has received intensive research attention over the past few years. Typically, the problem is investigated in the context of image classification, where addressing catastrophic forgetting effectively and efficiently is its core issue. 
We can broadly classify existing efforts into three categories: regularization~\cite{kirkpatrick2017ewc, Zenke2017CLTSI, Dhar2019LWM}, rehearsal~\cite{Parisi2019CLreview, Castro2018e2e, Aljundi2018MAS} and dynamic expansion~\cite{Aljundi2017expert_gate, Yan2021DER, Douillard2022Dytox, huang2023resolving}. Regularization methods emphasize constraining weight changes, e.g., allowing only small magnitude changes from the previous weights. It suffers from the problem that the changes do not adequately describe the complex pattern shifts caused by new task learning. Rehearsal methods keep a small amount of old data when training a new task, thus retaining some prior knowledge. Studies in this category focus on the selection of old data and the way it is used. For example, iCaRL was developed to learn an exemplar-based data representation~\cite{Rebuffi2017icarl}. Alternatively, dynamic expansion methods dynamically create feature extraction sub-networks each associated with one specific task~\cite{Golkar2019CLNR, Collier2020RNCCL, Wen2020Batchensemble, huang2023resolving}. Early methods required a task identifier to select the correct sub-network at test time. Unfortunately, the assumption is unrealistic as new samples would not come with their task identifiers. Recently, DER~\cite{Yan2021DER} proposed a dynamically expandable representation by discarding the task identifier, where the classifier was finetuned on a balanced exemplar subset to mitigate the task-tendency bias. It attained impressive results. Some recent works \cite{belouadah2019IL2M,delange2022clsurvey} studied IL in inhomogeneous or uneven datasets. However, the datasets they adopted are still ideal and cannot sufficiently describe challenges in real-world problems. Moreover, there were also some studies proposed for object detection \cite{feng2022overcoming, chen2019new, yang2022multi, yang2023one}, semantic segmentation \cite{yang2022uncertainty,douillard2021plop,zhang2022rcil} and object retrieval \cite{liu2023balanced}. Text recognition has not been studied in IL so far.

\subsection{Scene Text Recognition (STR)}
Text recognition is a longstanding research topic in computer vision and pattern recognition. Recent efforts mainly focused on recognizing text in natural scenes, i.e., STR. The task exhibits variations like text distortion, occlusion, blurring, etc., making the recognition challenging. With the advances in deep learning, especially CNN \cite{bai2014chinese,ShiBY17crnn,hu2020gtc,shi2018aster} and Transformers \cite{sheng2019nrtr,ABInet21CVPR,zheng2021cdistnet,wang2021visionlan,du2022@svtr,wang2022petr}, STR methods have been pushed forward significantly. 


Multilingual text recognition (MLTR) is an important sub-field of STR. The most popular solution for MLTR was data-joint training \cite{nayef2017mlt2017,buvsta2018e2emlt,nayef2019mlt2019,baek2020crafts}, where all data was gathered to train a model capable of recognizing all character classes. However, in addition to computational intensive, the approach also had the drawback of being biased toward data-rich languages, while performing poorly in minority languages where training data was scarce. As alternatives, multi-task or ensemble architectures were developed to allow data-rich languages to transfer knowledge to data-poor ones \cite{bai2014image,cui2017multilingual}. They alleviated the data scarcity issue to some extent. In addition, Some studies \cite{fujii2017seqscript,gomez2017improving,shi2015automatic,huang2021multiocr} added a script identification step to text recognition. They first identified the language domain and then selected the corresponding recognizer. Although similar to ours in the pipeline, they did not explore dependencies between languages. Moreover, none of them discussed the task within the IL framework.

\section{Methodology}
\subsection{Incremental Multilingual Text Recognition}
Our goal is to develop a unified model that can recognize text instances in different languages, with the model trained incrementally language-by-language. Mathematically, assume there are $I$ kinds of languages $\left\{\mathcal{D}_{1}, \cdots, \mathcal{D}_{I}\right\}$, with $\mathcal{D}_{i}=\left\{\left(\mathbf{x}_{i,1}, y_{i,1}\right), \cdots,\left(\mathbf{x}_{i,N(i)}, y_{i,N(i)}\right)\right\}$ as the training data at step $i$ (i.e., task $i$), where ${x}_{i,j}$ is the $j$-th input image and $y_{i, j} \in \mathcal{C}_{i}$ is its label within the label set $\mathcal{C}_{i}$, $N(i)$ is the number of samples in set $\mathcal{D}_{i}$. At the $i$-th learning step, samples of the $i$-th language will be added to the training set. Therefore, the goal can be formulated as to learn new knowledge from the set $\mathcal{D}_{i}$, while retaining the previous knowledge learned from old data $\left\{\mathcal{D}_{1}, \cdots, \mathcal{D}_{i-1}\right\}$. The label space of the model is all seen categories $\tilde{\mathcal{C}}_{i}=\cup_{k=1}^{i} \mathcal{C}_{k}$ and the model is expected to predict well on all classes in $\tilde{\mathcal{C}}_{i}$. Note that there may be a small overlap between label sets, i.e., $\mathcal{C}_{k} \cap \mathcal{C}_{j} \neq\emptyset$ for some $k$ and $j$. To better fight against catastrophic forgetting, we discuss IMLTR in the rehearsal setting. That is, a small and fix-sized rehearsal set $\mathcal{M}_i$ with a portion of samples from $\left\{\mathcal{D}_{1}, \cdots, \mathcal{D}_{i-1}\right\}$ is accessible at incremental step $i$. 


\subsection{Challenge and Solution Statement}

To build a recognition model to correctly recognize text instances from all currently seen languages and their character classes, let $x_{n}$ be the text instance to be recognized. $y_{n}^{t} \in \tilde{\mathcal{C}}_{i}$ denotes the $t$-th character label corresponding to $x_{n}$. $T(n)$ gives the total number of characters in this instance. 
IMLTR differs significantly from existing IL settings. For example compared to incremental image classification, standard IL usually has $|\tilde{\mathcal{C}}_{i}| \leq 100$ and $T(n)=1$ regardless of the value $n$. While the size of rehearsal set $\mathcal{M}_i$ is a constant (e.g., 2,000). However, in IMLTR $\mathcal{C}_{i}$ ranges from dozens of to thousands of character classes for different languages, and $T(n)$ belongs to (1, 25), assuming 25 as the maximized length of a character sequence.
Consequently, rehearsal-imbalance becomes a prominent challenge. Due to the limited size of the rehearsal set, it is not rare that a character class appears in the full dataset but is absent from the rehearsal set, as shown in Fig.~\ref{fig2:rehearsal-imbalance}. Thus, the incrementally trained models are likely to forget the absent character classes, despite having learned them previously, which can ultimately hurt the recognition accuracy.


Although the rehearsal set may not be enough to train a multilingual text recognizer to identify thousands of character classes, it is still sufficient to train a language classifier to recognize the language domains present in the text instance, whose classes are a much smaller number. Once the language domains are identified, we can choose an alternative scheme that involves aggregating the results from corresponding language recognizers to perform the recognition task, thereby bypassing the rehearsal-imbalance issue.

Motivated by this, we define $\mathcal{H}_{i}$ and $\varphi_{i}$ the skeleton network (all except classifier) and classifier trained at the $i$-th incremental step. Note that $\mathcal{H}_{i}$ is trained on $\mathcal{D}_{i}$, therefore can only recognize character classes of the $i$-th language in principle. Meanwhile, $\varphi_{i}$ is set to have $\tilde{\mathcal{C}}_{i}$ nodes to be compatible with typical IL settings, despite not being taught to recognize character classes of other languages. Then, we can adopt an aggregating-like scheme to implement IMLTR. The learning function can be written as:

\begin{equation}
\sum_{k=1}^{i}\prod_{t=1}^{T(n)}\left(P\left(y_{n}^{t} | x_{n} ; \mathcal{H}_{k}, \varphi_{k}\right) * S\left(d_{n}^{k} \right)\right),
\label{equ:equ2}
\end{equation}
where $d_{n}^{k}$ is the domain score indicating $x_{n}$ being classified as the \emph{k}-th language. $S(\cdot)$ is the score quantization function, which can be a one-hot vector (hard-voting) or a likelihood distribution (soft-voting). Eq.~\ref{equ:equ2} treats IMLTR as a weighted ensemble of recognition models trained based on different languages. By doing so, it successfully overcomes the rehearsal-imbalance issue within the IL framework. 

\subsection{Method Overview}
We propose a Multiplexed Routing Network (MRN) to implement this idea. As illustrated by Fig.~\ref{fig:overview}, it contains two stages, i.e., individual language modeling (stage-I) and multi-lingual modeling (stage-II). In stage-I, given $\mathcal{D}_{i}$ for the \emph{i}-th language, we train its recognizer using a popular text recognition model, which can recognize the character classes seen in $\mathcal{D}_{i}$. The model is represented as $\mathcal{H}_{i}$ and $\mathcal\varphi_{i}$. For character classes in $\tilde{\mathcal{C}}_{i}$ but not in $\mathcal{C}_{i}$, we simply truncate gradient propagation from these nodes thus the learned model still focuses on recognizing the $i$-th language.

Stage-II aims at building a multilingual routing network for IMLTR. Given a text instance $x_{n} \in \mathcal{D}_{i}\cup \mathcal{M}_i$, we feed it into all the learned \emph{i} skeleton networks in parallel, while keeping the parameters of the networks frozen for targeted feature extraction. It extracts \emph{i} sets of features, each associated with a certain language. The features are further fed into a Domain MLP Router (DM-Router) module, which is designed for domain score estimation, i.e., estimating the likelihood that the text instance belongs to the languages. Meanwhile, the \emph{i} sets of features are fed to their respective classifiers, where the corresponding recognition character sequences are obtained. To merge their recognition, we pad the classification nodes with zeros to $|\tilde{\mathcal{C}}_{i}|$, ensuring that all classifiers are aligned to the same dimension. As a result, their recognized characters can be merged using weighted element-wise addition, where the weights are the domain scores estimated using DM-Router. Finally, the recognition is conducted by applying a CTC- or attention-based decoding. Since DM-Router plays a critical role in the proposed method, we provide a detailed illustration below.

\subsection{Domain MLP Router}

\begin{figure}[]
\centering
\includegraphics[width=0.48\textwidth]{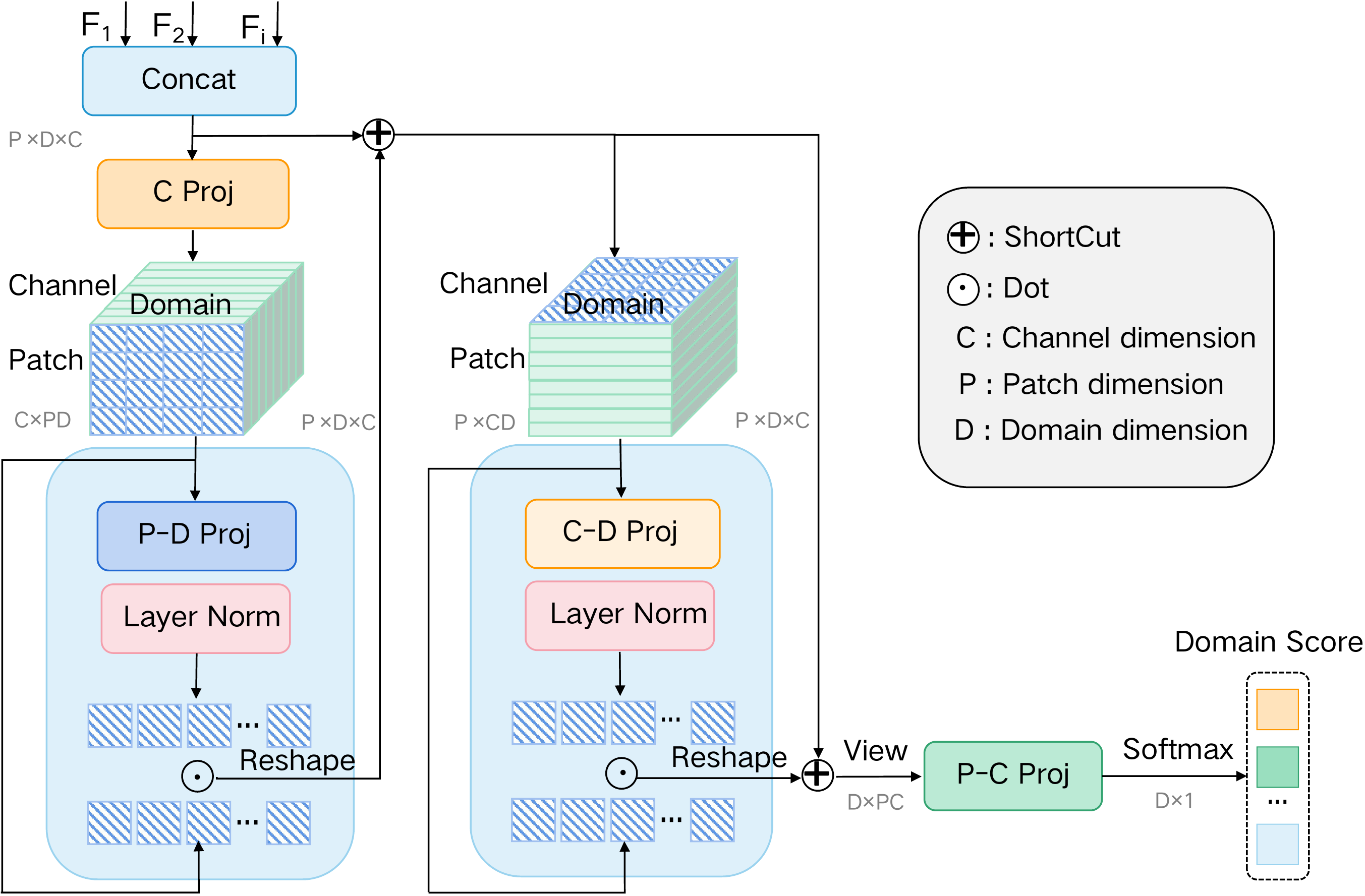} 
\caption{Detail structure of Domain MLP Router (DM-Router). Spatial-domain and Channel-domain dependencies are explored and fused to generate the language domain score distribution. }
\label{fig1:router}
\end{figure}

DM-Router uses features that are biased towards different language domains to discriminate the language domain of text instances. It accomplishes this by leveraging both the rehearsal set and the language data that arrives at the $i$-th step. While training a separate single-network classifier, which takes an image as input and outputs the language domain scores, can identify the language domains, we believe that this approach overlooks inter-domain dependencies that could be explored for better identification. For instance, different languages may have distinct appearance patterns, such as strokes, which differ significantly between Eastern Asian languages and Latin. Additionally, their features exhibit different frequency distributions, which can also aid language identification.

To achieve this goal, DM-Router accepts all \emph{i} sets of features extracted previously as input and mines the spatial-domain and channel-domain dependencies for better language identification. The detailed structure of DM-Router is shown in Fig.~\ref{fig1:router}. Features from different skeleton networks are concatenated, generating a feature cubic with size $P\times D \times C$, where $P$, $D$, and $C$ stand for the number of reshaped spatial patches, language domains, and feature channels, respectively. Then, a linear projection is applied along the channel dimension (C Proj), followed by reshaping the cubic from the patch-domain dimension. Next, a gated-mechanism is followed to generate the attention scores between the patch and domain. This is achieved by applying linear projection along the patch-domain dimension, followed by a layer norm and a feature dot product. We reshape the generated feature back to a feature cubic of the same size and merge it with the original cubic. The operations above explore the spatial-domain dependency. A similar operation is then applied to the merged feature cubic to explore the channel-domain dependence. In the following, the explored feature cubic gradually shrinks to a $D$-dimensional score vector that indicates the probability over the languages. It represents the likelihood of each language domain for the input text instance.   

DM-Router is an MLP-based attention network that targets language domain weighting. Note that there are a few similar solutions in the literature. Expert Gate (E-Gate) \cite{Aljundi2017expert_gate} developed an expert gating network that identified which model could be employed based on image reconstruction loss. However, it might not effectively discriminate IMLTR due to some languages exhibiting character class overlapping, which can cause classification confusion. On the other hand, multilingual OCR \cite{huang2021multiocr} determines the languages by script recognition and selected the corresponding model for recognition. Unlike these hard-voting methods, MRN employs soft-voting, which allows for the use of knowledge from other languages. For instance, Japanese has the ability to correct Chinese to some extent, given that they share some common words and similar strokes.

\subsection{Training Loss}
MRN has two loss terms. One for multilingual text recognition while the other for language domain prediction. The total loss function is written as:
\begin{equation}
\mathcal{L}_{\mathrm{total}}= \mathcal{L}_{\mathrm{clf}}+\alpha \mathcal{L}_{\mathrm{domain}},
\label{e5}
\end{equation}
where $\alpha$ is an empirical hyperparameter to balance the two.

MRN shows two advantages in dealing with rehearsal-imbalance. First, it ensures fair use of language. As previously mentioned, data distribution is uneven across different languages, and if not addressed during model training, it may lead to bias in the resulting model. By adopting language-by-language training and parameter freezing, data-rich and data-poor languages are treated equally, and class-imbalance is also alleviated. Second, MRN makes use of inter-lingual dependency in two ways: through the DM-Router described earlier, and through recognition score merging. When a character is recognized by more than one language, it receives confidence scores from each of them, allowing for the utilization of inter-lingual dependencies.

\begin{table}[]
\begin{center}
\resizebox{1.0\linewidth}{!}{
\begin{tabular}{c|c|cccccc}
\hline
\multirow{2}{*}{Dataset} & \multirow{2}{*}{categories} & Task1   & Task2 & Task3    & Task4  & Task5  & Task6  \\ \cline{3-8} 
                         &                       & Chinese & Latin & Japanese & Korean & Arabic & Bangla \\ \hline
\multirow{3}{*}{MLT17\cite{nayef2017mlt2017}} & train instance          & 2687 & 47411 & 4609 & 5631 & 3711 & 3237   \\
                         & test instance           & 529  & 11073 & 1350 & 1230 & 983  & 713    \\
                         & train class         & 1895 & 325   & 1620 & 1124 & 73   & 112    \\ \hline
\multirow{3}{*}{MLT19\cite{nayef2019mlt2019}} & train instance          & 2897 & 52921 & 5324 & 6107 & 4230 & 3542   \\
                         & test instance           & 322  & 5882  & 590  & 679  & 470  & 393    \\
                         & train class        & 2086 & 220   & 1728 & 1160 & 73   & 102    \\ \hline
\end{tabular}
}   
\end{center}
\caption{MLT17 and MLT19 statistics in our experiments.}
\label{table:mlt_datasets}
\end{table}

\section{Experiments}
\subsection{Datasets and Implementation Details} \label{DID}

\noindent\textbf{ICDAR 2017 MLT (MLT17) \cite{nayef2017mlt2017}}
has 68,613 training instances and 16,255 validation instances, which are from 6 scripts and 9 languages: Chinese, Japanese, Korean, Bangla, Arabic, Italian, English, French, and German. The last four use Latin script. The samples are from natural scenes with challenges like blur, occlusion, and distortion. We use the validation set for test due to the unavailability of test data. Tasks are split by scripts and modeled sequentially. Special symbols are discarded at the preprocessing step as with no linguistic meaning.

\begin{table*}[]
\begin{center}
\resizebox{0.95\linewidth}{!}{
\begin{tabular}{ccccccccccccccc}

\hline
\multicolumn{1}{l|}{}         & \multicolumn{7}{c|}{MLT17}                                                                                        & \multicolumn{7}{c}{MLT19}                                                                    \\ \hline
\multicolumn{10}{l}{Model : CRNN (TPAMI'17) \cite{ShiBY17crnn}}                                                                                                                  \\ \hline
\multicolumn{1}{c|}{ }     & T1   & T2            & T3            & T4            & T5            & T6            & \multicolumn{1}{c|}{AVG}           & T1   & T2            & T3            & T4            & T5            & T6            & AVG           \\ \hline
\multicolumn{1}{c|}{Bound}    & -    & -             & -             & -             & -             & -             & \multicolumn{1}{c|}{92.1}          & -    & -             & -             & -             & -             & -             & 84.9          \\
\multicolumn{1}{c|}{Baseline} & 91.1 & 51.7          & 51.0          & 37.2          & 29.3          & 22.3          & \multicolumn{1}{c|}{47.1}          & 85.1 & 49.6          & 46.5          & 35.5          & 27.6          & 20.7          & 44.2          \\ \hline
\multicolumn{1}{c|}{LwF (TPAMI'17)\cite{Li2017LWF}}      & 91.1 & 53.7          & 53.4          & 38.2          & 29.7          & 23.7          & \multicolumn{1}{c|}{48.3}          & 85.1 & 51.6          & 49.2          & 36.5          & 27.7          & 22.0          & 45.3          \\
\multicolumn{1}{c|}{EWC (PNAS'17)\cite{kirkpatrick2017ewc}}      & 91.1 & 56.5          & 50.4          & 37.2          & 30.5          & 21.5          & \multicolumn{1}{c|}{47.9}          & 85.1 & 55.5          & 46.3          & 35.8          & 28.8          & 19.9          & 45.2          \\
\multicolumn{1}{c|}{WA  (CVPR'20) \cite{zhao2020wa}}       & 91.1 & 54.6          & 48.7          & 38.2          & 28.5          & 23.1          & \multicolumn{1}{c|}{47.4}          & 85.1 & 52.2          & 44.3          & 36.7          & 26.8          & 21.6          & 44.4          \\
\multicolumn{1}{c|}{DER (CVPR'21)\cite{Yan2021DER}}      & 91.1 & 76.3          & 55.8          & 46.4          & 39.3          & 35.8          & \multicolumn{1}{c|}{57.5}          & 85.1 & 75.2          & 40.4          & 45.1          & 36.6          & 34.2          & 52.8          \\ \hline
\multicolumn{1}{c|}{MRN}     & 91.1 & \textbf{88.6} & \textbf{77.2} & \textbf{73.7} & \textbf{69.8} & \textbf{69.8} & \multicolumn{1}{c|}{\textbf{78.4}} & 85.1 & \textbf{85.1} & \textbf{73.2} & \textbf{68.3} & \textbf{65.3} & \textbf{65.5} & \textbf{73.7} \\ \hline
\hline
\multicolumn{15}{l}{Model : TRBA (ICCV'19) \cite{Baekwhats_wrong_19ICCV} }                                                                                      \\ \hline
\multicolumn{1}{c|}{}         & T1   & T2            & T3            & T4            & T5            & T6            & \multicolumn{1}{c|}{AVG}           & T1   & T2            & T3            & T4            & T5            & T6            & AVG           \\ \hline
\multicolumn{1}{c|}{Bound}    & -    & -             & -             & -             & -             & -             & \multicolumn{1}{c|}{94.9}          & -    & -             & -             & -             & -             & -             & 90.5          \\
\multicolumn{1}{c|}{Baseline} & 91.3 & 49.6          & 47.3          & 36.1          & 28.6          & 24.0          & \multicolumn{1}{c|}{46.1}          & 85.4 & 49.4          & 44.0          & 34.8          & 27.4          & 23.1          & 44.0          \\ \hline
\multicolumn{1}{c|}{LwF (TPAMI'17)\cite{Li2017LWF}}      & 91.3 & 55.7          & 38.8          & 28.7          & 22.6          & 18.7          & \multicolumn{1}{c|}{42.6}          & 85.4 & 54.2          & 35.0          & 27.2          & 20.5          & 17.0          & 39.9          \\
\multicolumn{1}{c|}{EWC (PNAS'17)\cite{kirkpatrick2017ewc}}      & 91.3 & 50.4          & 43.6          & 33.1          & 25.6          & 21.9          & \multicolumn{1}{c|}{44.3}          & 85.4 & 49.4          & 40.6          & 31.7          & 24.8          & 20.6          & 42.1          \\
\multicolumn{1}{c|}{WA (CVPR'20) \cite{zhao2020wa}}       & 91.3 & 45.4          & 41.8          & 30.7          & 23.5          & 19.6          & \multicolumn{1}{c|}{42.1}          & 85.4 & 44.0          & 37.9          & 29.2          & 21.6          & 18.1          & 39.4          \\
\multicolumn{1}{c|}{DER (CVPR'21)\cite{Yan2021DER}}      & 91.3 & 60.1          & 53.0          & 38.8          & 31.4          & 28.6          & \multicolumn{1}{c|}{50.5}          & 85.4 & 60.7          & 50.3          & 37.2          & 30.3          & 28.1          & 48.7          \\ \hline
\multicolumn{1}{c|}{MRN}     & 91.3 & \textbf{87.9} & \textbf{75.8} & \textbf{72.2} & \textbf{71.5} & \textbf{68.7} & \multicolumn{1}{c|}{\textbf{77.9}} & 85.4 & \textbf{84.5} & \textbf{73.2} & \textbf{67.8} & \textbf{66.7} & \textbf{64.8} & \textbf{73.7} \\ \hline
\hline
\multicolumn{15}{l}{Model : SVTR-Base (IJCAI'22) \cite{du2022@svtr}}                                         \\ \hline
\multicolumn{1}{c|}{}         & T1   & T2            & T3            & T4            & T5            & T6            & \multicolumn{1}{c|}{AVG}           & T1   & T2            & T3            & T4            & T5            & T6            & AVG           \\ \hline
\multicolumn{1}{c|}{Bound}    & -    & -             & -             & -             & -             & -             & \multicolumn{1}{c|}{90.1}          & -    & -             & -             & -             & -             & -             & 83.2          \\
\multicolumn{1}{c|}{Baseline} & 90.6 & 32.5          & 40.5          & 30.8          & 24.5          & 19.9          & \multicolumn{1}{c|}{39.8}          & 84.8 & 31.3          & 37.0          & 29.2          & 22.6          & 19.1          & 37.3          \\ \hline
\multicolumn{1}{c|}{LwF (TPAMI'17)\cite{Li2017LWF}}      & 90.6 & 28.0          & 38.4          & 29.9          & 24.1          & 18.3          & \multicolumn{1}{c|}{38.2}          & 84.8 & 27.0          & 34.6          & 28.4          & 22.3          & 17.0          & 35.7          \\
\multicolumn{1}{c|}{EWC (PNAS'17)\cite{kirkpatrick2017ewc}}      & 90.6 & 33.0          & 41.2          & 31.1          & 24.6          & 20.0          & \multicolumn{1}{c|}{40.1}          & 84.8 & 31.3          & 37.7          & 29.5          & 22.6          & 19.0          & 37.5          \\
\multicolumn{1}{c|}{WA  (CVPR'20) \cite{zhao2020wa}}       & 90.6 & 28.0          & 37.9          & 30.4          & 24.8          & 19.8          & \multicolumn{1}{c|}{38.6}          & 84.8 & 26.7          & 34.6          & 28.3          & 22.6          & 18.6          & 35.9          \\
\multicolumn{1}{c|}{DER (CVPR'21)\cite{Yan2021DER}}      & 90.6 & 74.5          & 55.7          & 55.0          & 49.5          & 45.7          & \multicolumn{1}{c|}{61.8}          & 84.8 & 71.6          & 52.9          & 52.2          & 46.6          & 43.6          & 58.6          \\ \hline
\multicolumn{1}{c|}{MRN}     & 90.6 & \textbf{86.4} & \textbf{73.9} & \textbf{65.6} & \textbf{63.4} & \textbf{58.1} & \multicolumn{1}{c|}{\textbf{73.0}} & 84.8 & \textbf{83.7} & \textbf{69.4} & \textbf{64.4} & \textbf{57.8} & \textbf{53.1} & \textbf{68.9} \\ \hline
\end{tabular}
}
\end{center}
\caption{Accuracy (\%) of different text recognizers and incremental learning methods on MLT17 and MLT19. \emph{Baseline} denotes the model trained solely based on the rehearsal set and language data arrived at that step. The language incremental order is introduced in Sec.~\ref{DID}.
}
\label{table:sota}
\end{table*}

\noindent\textbf{ICDAR 2019 MLT (MLT19) \cite{nayef2019mlt2019}} has 89,177 text instances coming from 7 scripts. Since the inaccessibility of test set, we randomly split the training instances to 9:1 script-by-script, for model training and test. To be consistent with MLT2017 dataset, we discard the Hindi script and also special symbols. Statistics of the two datasets are shown Tab.~\ref{table:mlt_datasets}.

Height and width of the images are scaled uniformly to $32\times256$. The maximum length of a character sequence is set to 25. All models, each corresponding to a language domain, are trained with 10,000 iterations, using the Adam optimizer and the one-cycle learning rate scheduler \cite{smith2019super} with a maximum learning rate of 0.0005. The batch size is set to 256. To mitigate the dataset variance, in each batch we evenly sample training samples from both datasets, that is, half from MLT17 and half from MLT19. A random order for the six languages is employed, which is Chinese, Latin, Japanese, Korean, Arabic, Bangla. Other orders will be discussed later. For the rehearsal setting, we limit the rehearsal size to 2000 samples unless specified. We conduct the experiments using two NVIDIA RTX 3090 GPUs.

\subsection{Comparison with Existing Methods}
We equip MRN with different text recognizers and combine them with different IL methods. Specifically, we consider three typical STR schemes: CTC-based (CRNN \cite{ShiBY17crnn}), attention-based (TRBA \cite{Baekwhats_wrong_19ICCV}), and ViT-based (SVTR \cite{du2022@svtr}). Meanwhile, four popular IL methods are chosen, i.e., Lwf \cite{Li2017LWF}, EWC \cite{kirkpatrick2017ewc}, WA \cite{zhao2020wa} and DER \cite{Yan2021DER}. All models retain their original settings, except for the removal of the auxiliary loss of DER, which reduces its performance in our task.

In Tab.~\ref{table:sota}, we give the results at different incremental steps, where the language is added one-by-one and the average accuracy of different methods is reported. \emph{Bound}, the model trained using all training data, is also listed as the oracle for reference. As can be seen, MRN consistently outperforms all the compared methods by significant margins under different settings, no matter which recognizer is employed. When looking into the general-purpose IL methods, their accuracy mostly decreased rapidly as the incremental steps due to the affection of rehearsal-imbalance. DER has the highest accuracy among them, as its dynamic expansion architecture has certain advantages in fighting against catastrophic forgetting. However, there is still a clear accuracy gap between DER and our MRN, and the gap widens as the incremental step increases. We attribute the accuracy improvement achieved by MRN to two factors. First, IMLTR is a task that differs significantly from image classification, where most IL methods have been experimented on. These methods do not well accommodate the challenge raised by IMLTR. For example, the rehearsal-imbalance issue. Second, MRN develops an elegant pipeline that implements the recognition in a domain routing and result fusion manner. It works particularly well for scenarios where incremental tasks exhibit significant differences. 

When comparing the recognizers, MRN equipped with CRNN has the highest overall accuracy. The result is interesting as CRNN has a simpler architecture and generally performs worse than the other two methods on typical STR tasks. We attribute this to parameter freezing, where the feature extraction backbone (e.g., $\mathcal{H}_{i}$) and the decoder cannot be jointly optimized. Therefore, advanced models are more severely affected, while the simpler one is less affected and can better mitigate catastrophic forgetting.

\begin{table}[h]
\begin{center}
\begin{tabular}{c|cc|cc}
\hline
\multirow{2}{*}{model} & \multicolumn{2}{c|}{MLT17}    & \multicolumn{2}{c}{MLT19}     \\ \cline{2-5} 
                       & Avg           & Last          & Avg           & Last          \\ \hline
None                   & 64.8          & 37.9          & 60.8          & 35.6          \\
MLP                 & 68.5          & 60.5          & 65.3          & 56.3          \\
CycleMLP\cite{chen2022cyclemlp}              & 75.5          & 63.5          & 71.1          & 60.0          \\
ViP\cite{hou2022vip}                    & 76.4          & 62.6          & 72.2          & 59.6          \\
gMLP\cite{liu2021gmlp}                   & 77.5          & 68.2          & 73.1          & 64.2          \\ \hline
DM-Router                   & \textbf{78.4} & \textbf{69.8} & \textbf{73.8} & \textbf{65.5} \\ \hline
\end{tabular}
\end{center}
\caption{
Performance comparisons on different MLP models.
}  
\label{table:mlp}
\end{table}

\subsection{Ablation Study}
We perform a series of controlled experiments to gain a deeper understanding of MRN. CRNN is employed as the text recognizer unless specified.

\noindent\textbf{Effectiveness of DM-Router:} There are multiple ways to deduce the language domain scores. We enumerate several of them that have been used in existing studies, as shown in Tab.~\ref{table:mlp}. \emph{None} denotes no dependence is explored, which corresponds to the worst result. It, in turn, demonstrates the necessity of utilizing language dependence. Among the rest competitors, MLP enables a naive learning mechanism while the remaining three are based on more advanced MLP-like models, which are typically more effective. Despite this, DM-Router attains the highest accuracy among the methods. The results clearly demonstrate the rationality of the DM-Router structure in terms of language dependence exploration.

\begin{table}[t]
\begin{center}
\resizebox{0.8\linewidth}{!}{
\begin{tabular}{c|c|cc|cc}
\hline
\multirow{2}{*}{Size} & \multirow{2}{*}{Method} & \multicolumn{2}{c|}{MLT17}                   & \multicolumn{2}{c}{MLT19}                   \\ \cline{3-6} 
                     &                         & Avg                  & Last                  & Avg                  & Last                 \\ \hline
\multirow{3}{*}{2k}  & LwF\cite{Li2017LWF}                     & 48.3                 & 23.7                  & 45.4                 & 22.0                 \\
                     & DER\cite{Yan2021DER}                     & 57.5                 & 35.8                  & 52.8                 & 34.2                 \\
                     & MRN                  &  \textbf{78.4} & \textbf{69.8} & \textbf{73.8} & \textbf{65.5}              \\ \hline
\multirow{3}{*}{3k}  & LwF\cite{Li2017LWF}                     & 52.2                 & 24.9                  & 48.8                 & 23.6                 \\
                     & DER\cite{Yan2021DER}                     & 60.9                 & 42.0                  & 58.7                 & 40.6                 \\
                     & MRN                   & \textbf{80.2} & \textbf{72.7} & \textbf{75.4} & \textbf{68.2}                \\ \hline
\multirow{3}{*}{4k}  & LwF\cite{Li2017LWF}                     & 55.5                 & 27.5                  & 52.2                 & 26.1                 \\
                     & DER\cite{Yan2021DER}                     & 66.4                 & 48.7                  & 63.8                 & 46.6                 \\
                     & MRN                    & \textbf{81.5} & \textbf{75.0} & \textbf{76.5} & \textbf{70.6}
                    \\ \hline
\end{tabular}}
\end{center}
\caption{Ablation study on the size of the rehearsal set.}
\label{tab:rec_pos}
\end{table}

\noindent\textbf{Influence of the size of the rehearsal set:} We conduct analytical experiments to evaluate the influence of the rehearsal size on the accuracy of Lwf \cite{Li2017LWF}, DER \cite{Yan2021DER} and MRN. Fig.~\ref{tab:rec_pos} shows the accuracy under different rehearsal sizes. As anticipated, increasing the rehearsal set size leads to accuracy gains, as more past memories are retained. We observe that larger gains are obtained in LwF and DER, particularly DER. This reveals the accuracy of general-purpose IL methods is largely affected by the rehearsal size in IMLTR, while MRN is less affected. MRN has already achieved relatively high accuracy, and the performance of MRN in identifying language domains is less affected by the rehearsal size. The results indicate that MRN is robust to rehearsal scarcity and can better fight against data imbalance.

\noindent\textbf{Influence of language incremental order:} In addition to the order in Sec.~\ref{DID} (O1), we assess two other orders as follows: 1) Arabic, Chinese, Latin, Japanese, Bangla, Korean (O2); 2) Latin, Arabic, Bangla, Chinese, Japanese, Korean (O3). The two orders either alternate the three Eastern Asia languages, which have large vocabularies and show more stroke commonalities, or group them together at the end. We also include Lwf and DER for comparison.
\begin{table}[]
\begin{center}
\begin{tabular}{c|c|cc|cc}
\hline
\multirow{2}{*}{Order} & \multirow{2}{*}{Method} & \multicolumn{2}{c|}{MLT17} & \multicolumn{2}{c}{MLT19} \\ \cline{3-6} 
                       &                         & Avg          & Last        & Avg         & Last        \\ \hline
\multirow{3}{*}{O1}    & LwF\cite{Li2017LWF}                     & 48.3         & 23.7        & 45.4        & 22.0        \\
                       & DER\cite{Yan2021DER}                     & 57.5         & 35.8        & 52.8        & 34.2        \\
                       & MRN                    &\textbf{78.4} & \textbf{69.8} & \textbf{73.8} & \textbf{65.5}  \\ \hline

\multirow{3}{*}{O2}    & LwF\cite{Li2017LWF}                     & 46.9         & 23.8        & 43.1        & 22.9        \\
                       & DER\cite{Yan2021DER}                     & 63.1         & 39.1        & 58.7        & 39.6        \\
                       & MRN                    & \textbf{80.5} & \textbf{65.3} & \textbf{74.1} & \textbf{61.5}  \\ \hline
\multirow{3}{*}{O3}    & LwF\cite{Li2017LWF}                     & 57.7         & 34.7        & 55.7        & 34.2        \\
                       & DER\cite{Yan2021DER}                     & 69.6         & 41.3        & 65.7        & 38.2        \\
                       & MRN                     & \textbf{82.9} & \textbf{70.6} & \textbf{78.3} & \textbf{66.0}\\ \hline                
\end{tabular}
\end{center}
\caption{Ablation study on language order.}
\label{table:order}
\end{table}

\begin{table}[]
\begin{center}
\begin{tabular}{c|cc|cc}
\hline
\multirow{2}{*}{Sampling Strategy} & \multicolumn{2}{c|}{MLT17}    & \multicolumn{2}{c}{MLT19}     \\ \cline{2-5} 
                                  & Avg           & Last          & Avg           & Last          \\ \hline
Confidence                           & 56.4          & 43.8          & 54.0          & 41.2          \\
Length                              & 71.0          & 50.3          & 66.6          & 48.9          \\Frequency                          & 72.6          & 56.6          & 67.8          & 53.7          \\ \hline
Random                             & \textbf{78.4} & \textbf{69.8} & \textbf{73.8} & \textbf{65.5} \\ \hline
\end{tabular}
\end{center}
\caption{Ablation study on rehearsal sampling strategy.}
\label{table:sampling}
\end{table}

\begin{table*}[]
\begin{center}
\resizebox{0.8\linewidth}{!}{
\begin{tabular}{c|ccc|cc|cc|c|c}
\hline
\multirow{2}{*}{Method} & \multirow{2}{*}{Select} & \multirow{2}{*}{Model} & \multirow{2}{*}{Voting} & \multicolumn{2}{c|}{MLT17}    & \multicolumn{2}{c|}{MLT19} &  \multirow{2}{*}{Params (M)} & \multirow{2}{*}{FLOPs (G)} \\ \cline{5-8} 
                        &                         &                        &                         & Avg           & Last          & Avg           & Last & &          \\ \hline
Baseline                    & --                      & --                     & --                      & 47.1          & 22.3          & 44.2          & 20.7 & 9.5 & 3.5       \\
DER\cite{Yan2021DER}                     & --                      & --                     & --                      & 57.5          & 35.8          & 52.8          & 34.2         & 33.8 & 12.3 \\ \hline
E-Gate\cite{Aljundi2017expert_gate}                  & Re-Const.          & Autoencoder            & Hard                    & 37.2          & 15.2          & 34.8          & 14.2     & 32.5 &     12.2 \\
E-Gate\cite{Aljundi2017expert_gate}                  & Stacking                & Autoencoder            & Hard                    & 62.7          & 15.2          & 59.3          & 14.2     & 35.5 & 12.4   \\ \hline
MRN                    & Stacking                & DM-Router               & Hard                    & 74.4          & 62.9          & 69.9          & 57.7  & 33.5 & 12.4       \\ 
MRN                    & Stacking                & DM-Router               & Soft                   & \textbf{78.4} & \textbf{69.8} & \textbf{73.8} & \textbf{65.5} & 33.5 & 12.4\\ \hline
\end{tabular}}
\end{center}
\caption{Comparisons on different routing strategies.}
\label{table:e-gate}
\end{table*}

Tab.~\ref{table:order} gives the results and two observations. First, O3 shows the best accuracy, while O2 also performs better than O1. It is because the three Eastern Asia languages are more difficult to recognize due to their large vocabulary sizes, therefore introducing them later leads to a better average accuracy. Meanwhile, putting them together also reduces the oscillation during parameter learning and generates a better model, due to their stroke commonalities. The experiment suggests that careful selection of the order of languages can attain better accuracy. Second, O1 shows the largest accuracy gaps between MRN and other methods. This is because in O1, the large vocabulary languages appear earlier, while the rehearsal set is fix-sized, resulting in the most severe class imbalance among the three orders. The result indicates that MRN can better handle class imbalance.

\noindent\textbf{Influence of rehearsal sampling strategy:} The determination of text instances being sampled to the rehearsal set is an issue also worthy of ablating. Tab.~\ref{table:sampling} gives the accuracy of four sampling strategies, i.e., \emph{Confidence} that selects instances with the highest recognition scores, \emph{Length} that selects instances with the largest number of characters, \emph{Frequency} that selects instances with the most frequently occurred characters, and \emph{Random} adopted in our MRN that randomly selects the instances. Interestingly, \emph{Random} gives the best accuracy. We attribute the reason to: the rehearsal set obtained from \emph{Confidence} or \emph{Frequency} cannot fully represent the true data distribution, where difficult or less occurred instances are excluded. \emph{Length}, to some extent, overlooks the varying-length characteristic of IMLTR. On the contrary, \emph{Random}, despite simple, well mimics the underlying data distribution and well handles the variable length challenge.

\noindent\textbf{Comparison on routing strategy:} We compare MRN with E-Gate and its variants. E-Gate \cite{Aljundi2017expert_gate} treats different sub-networks as experts, and each time selects the most appropriate one for inference. In Tab.~\ref{table:e-gate} we provide the model details. Raw E-Gate performs poorly in IMLTR. When stacking is used to build feature extractors, the accuracy improves significantly and outperforms DER. We also evaluate MRN with hard-voting. It reports a worse result. Compared to other routing strategies, our MRN shows clear superiority in terms of accuracy, while incurring only a negligible cost in parameters and computational complexity.

\begin{figure}[]
\centering
\includegraphics[width=0.45\textwidth,height=0.24\textheight]{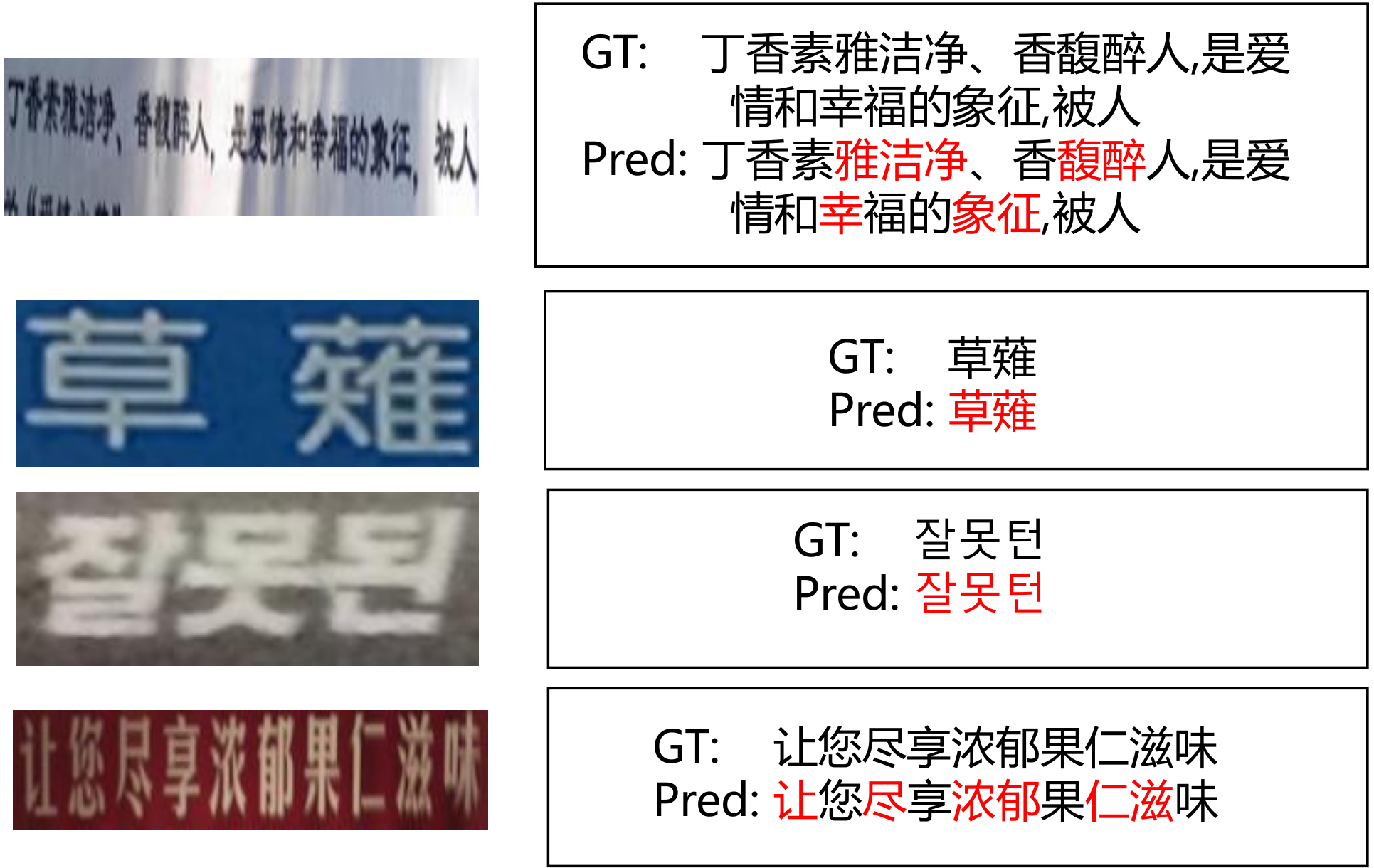}
\caption{Illustrative recognition examples, where \textcolor{red}{red} denotes characters that are absent from the rehearsal set.}
\label{fig:unclass_vis}
\end{figure}

\subsection{Qualitative Results Analysis}
Fig.~\ref{fig:unclass_vis} gives several recognition results of MRN. It correctly read instances of different languages, even with the presence of common recognition difficulties. More importantly, MRN also recognizes character classes that are not present in the rehearsal set. These results again demonstrate that our MRN is effective in handling rehearsal-imbalance and can generalize well to unseen character classes.


\section{Conclusion}
In this work, we introduce a new task called incremental multilingual text recognition (IMLTR). IMLTR handles text recognition in an incremental learning setting, therefore is suitable for applications like streaming data processing. IMLTR faces a distinct problem of rehearsal-imbalance, including data imbalance, class imbalance, and variable character length. To address this challenge, we designed a Multiplexed Routing Network (MRN) that first trains a multi-language correlated DM-router to weight the language domains, and then votes the separately trained recognition branches for final text recognition. Experiments on public benchmarks show that MRN significantly outperforms existing general-purpose IL methods by large margins. As the first attempt to apply IL to multilingual text recognition, we hope that this work will broaden the applications of text recognition and inspire further research in this area.

\vspace{0.3cm}
\noindent\textbf{Acknowledgments}
This project was supported by National Key R\&D Program of China (No. 2022YFB3104703) and in part by the National Natural Science Foundation of China (No. 62172103)

{\small
\bibliographystyle{ieee_fullname}
\bibliography{egbib}
}

\newpage

\ificcvfinal\thispagestyle{empty}\fi


\maketitle
\appendix

\section{Analyses on the Domain Loss Weights}
The first is the $\alpha$ in manuscript Equ.3, which adjusts the contribution of language domain identification and text recognition. Tab.~\ref{table:para_a} shows the results on several enumerated $\alpha$. The accuracy experiences a firstly increased and then decreased procedure, where the best accuracy is reached when $\alpha$ equals to 15. Text recognition loss plays a dominant role in MRN. This is reasonable as text recognition is the main task while it accumulates loss at the character level. This also indicates that the classification of language domains also contributes to joint optimization. 

\begin{table}[]
\begin{center}
\begin{tabular}{c|cc|cc}
\hline
\multirow{2}{*}{$\alpha$} & \multicolumn{2}{c|}{MLT17}    & \multicolumn{2}{c}{MLT19}     \\ \cline{2-5} 
                           & Avg           & Last          & Avg           & Last          \\ \hline
1                          & 75.4          & 66.5          & 71.5          & 64.4          \\
5                          & 76.3          & 68.2          & 71.9          & 64.0          \\
10                         & 76.8          & 68.1          & 72.1          & 62.7          \\
15                         & \textbf{78.4} & \textbf{69.8} & \textbf{73.8} & \textbf{65.5} \\
20                         & 77.5          & 67.4          & 72.8          & 64.3          \\ \hline
\end{tabular}
\end{center}
\caption{Ablation study on domain loss weights ($\alpha$).}
\label{table:para_a}
\end{table}

\section{Analyses on the Number of DM-Router}

In DM-Router module, one might guess that stacking the internal language dependence exploration parts several times may drive a better dependence utilization. We also empirically validate this and Tab.\ref{table:num_dm} gives the results. As seen, the results answer that performing DM-Router module once is sufficient to explore the language dependence. It in turn proves the rationality of the DM-Router structure.
\begin{table}[]
\begin{center}
\begin{tabular}{c|cc|cc}
\hline
\multirow{2}{*}{\# Router} & \multicolumn{2}{c|}{MLT17}    & \multicolumn{2}{c}{MLT19}     \\ \cline{2-5} 
                         & Avg           & Last          & Avg           & Last          \\ \hline
1                         & \textbf{78.4} & \textbf{69.8} & \textbf{73.8} & \textbf{65.5} \\
2                         & 77.1          & 68.3          & 72.5          & 64.2          \\
3                         & 76.8          & 68.1          & 72.1          & 62.7          \\
\hline       
\end{tabular}
\end{center}
\caption{Ablation study on the number of DM-Router modules.}
\label{table:num_dm}
\end{table}

\begin{figure}[htb]
\centering
\includegraphics[width=0.45\textwidth]{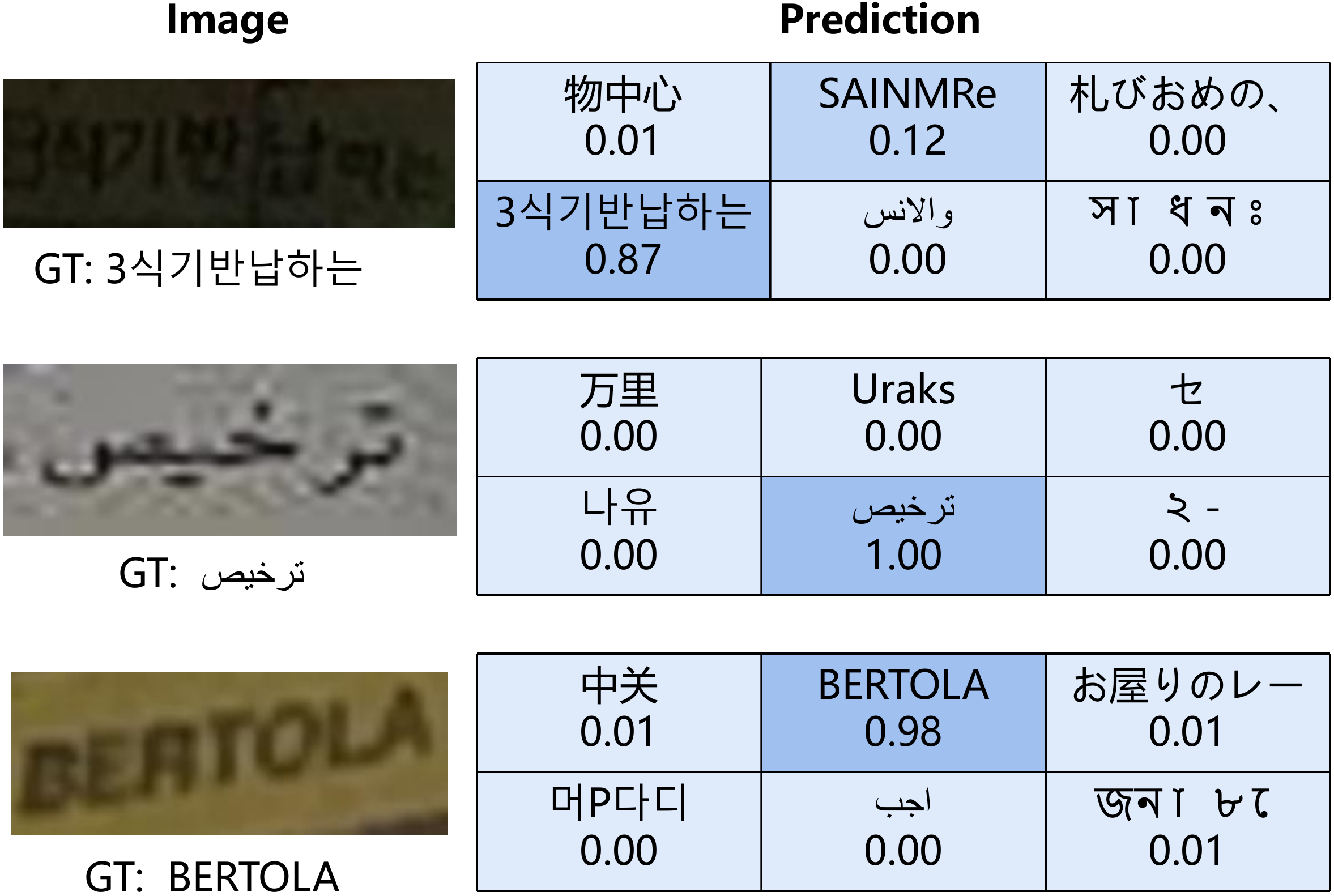} 
\caption{Illustrative examples with their predictions and domain scores.}
\label{fig:expert_vis}
\end{figure}
\section{Visualizes on the DM-Router}
In Fig.~\ref{fig:expert_vis}, we visualize three examples, each containing a different language. DM-Router gives considerable well language identification, and each language all gives its recognition.

\begin{figure*}[thbp]
\centering
\includegraphics[width=1.0\textwidth]{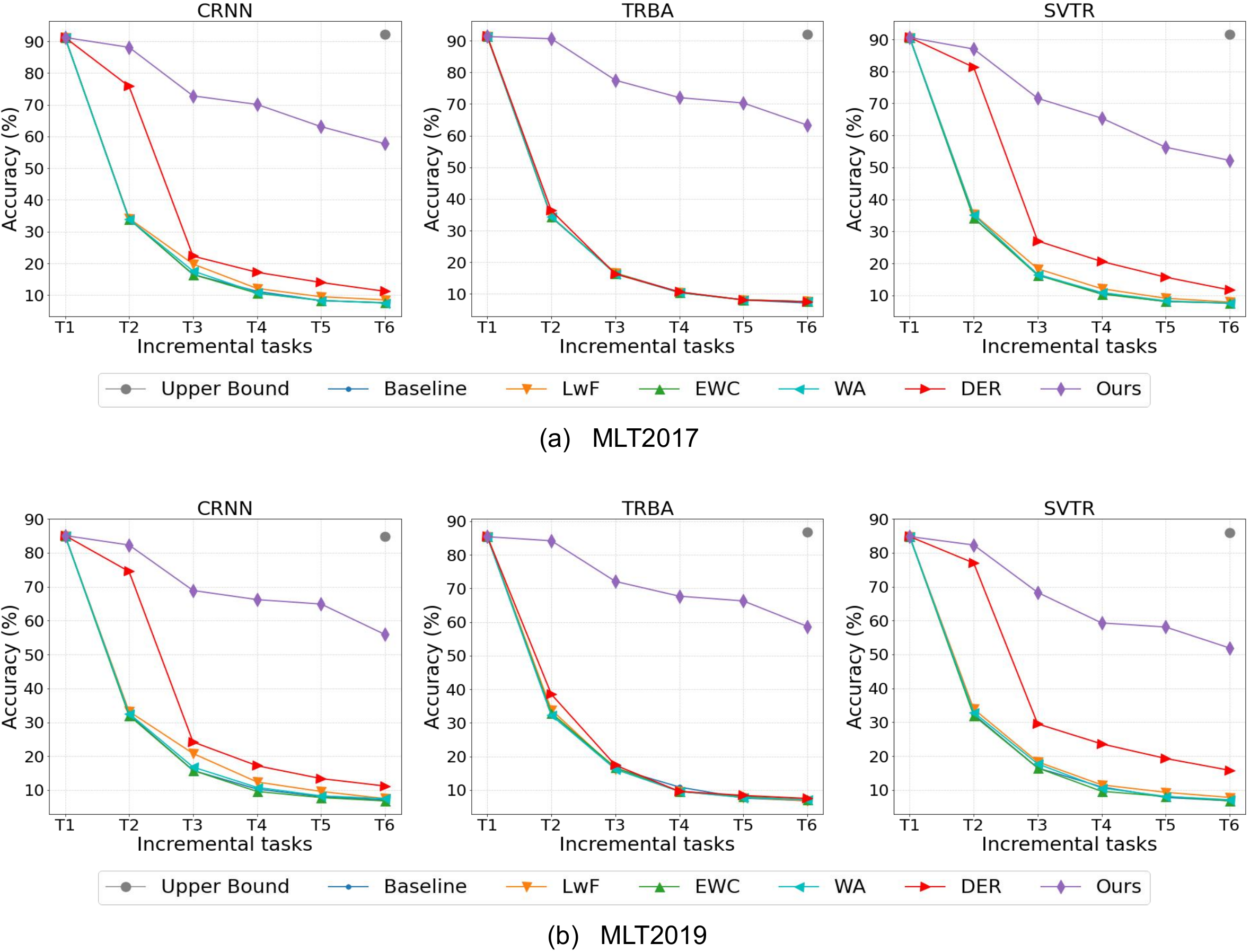} 
\caption{The Accuracy (\%) of Chinese data on MLT17 and MLT19 tested against different combinations of text recognition and incremental learning methods.}
\label{fig:chinese acc}
\end{figure*}
\section{Accuracy of Chinese Data at Each Task}

To analyze the degree of forgetting the same language by different incremental methods, we show the accuracy of Chinese data tested on MLT17 and MLT19 for different combinations of text recognition and incremental learning methods.MLT17 and MLT19 show the same trend. Incremental learning methods (except MRN and DER) are completely unable to maintain the memory of the old language due to the rehearsal-imbalance. DER memorability depends on a larger memory budget, as the memory budget decreases, DER is unable to maintain the memorability of the old language. MRN maintains a stable performance advantage and is less dependent on the rehearsal memory budget.


\end{document}